\documentclass{article}
\usepackage{spconf}
\usepackage{amssymb}
\usepackage{amsmath}
\usepackage{graphicx}
\usepackage{lipsum}
\usepackage{dblfloatfix}
\usepackage{hyperref}

\title{Large Vision Models can solve mental rotation problems}
\newcommand{\secondlinegap}{1.5em} 

\name{%
  \begin{tabular}{c}
    Sebastian Ray Mason\textsuperscript{*}\hspace{\secondlinegap}
    Anders Gj{\o}lbye\textsuperscript{*}\textsuperscript{\textdagger}\hspace{\secondlinegap}
    Phillip Chavarria H{\o}jbjerg \\ [0.5em]
    \textit{Lenka T\v{e}tkov\'a}\hspace{\secondlinegap}
    \textit{Lars Kai Hansen}
  \end{tabular}
  \thanks{\textsuperscript{*} These authors contributed equally as first authors.}%
  \thanks{\textsuperscript{\dag} Corresponding author.}
}

\address{
Technical University of Denmark\\
Section for Cognitive Systems\\
2800 Kgs.\ Lyngby, Denmark\\ [1ex]
\texttt{sebastianraymason@outlook.com}\qquad \texttt{\{agjma, pchho, lenhy, lkai\}@dtu.dk}\\[-0.5em]}

\ninept
\begin{document}
\maketitle
\begin{abstract}
Mental rotation is a key test of spatial reasoning in humans and has been central to understanding how perception supports cognition. Despite the success of modern vision transformers, it is still unclear how well these models develop similar abilities. In this work, we present a systematic evaluation of \texttt{ViT}, \texttt{CLIP}, \texttt{DINOv2}, and \texttt{DINOv3} across a range of mental-rotation tasks, from simple block structures similar to those used by Shepard and Metzler to study human cognition, to more complex block figures, three types of text, and photo-realistic objects. By probing model representations layer by layer, we examine where and how these networks succeed. We find that i) self-supervised ViTs capture geometric structure better than supervised ViTs; ii) intermediate layers perform better than final layers; iii) task difficulty increases with rotation complexity and occlusion, mirroring human reaction times and suggesting similar constraints in embedding space representations.
\end{abstract}



\section{Introduction}
Modern self-supervised representation learning is scoped to capture the full data distribution, including concept semantics and context dimensions. For example, general vision models should represent foreground objects as well as their context, such as background, location, pose, color, illumination, etc. While detection of concept semantics typically leads to transformation {\it invariant} models, prediction of context variables is promoted by {\it equivariant} representations as shown in Cosentino et al. \cite{cosentino2022geometry}.

In the interest of AI alignment, we investigate the invariance/equivariance of machine learning models and our current understanding of human cognition. Human representations are only indirectly accessible through, e.g., behavioral studies or brain imaging. Evidently, humans can task themselves to focus on both object semantics or context variables at will. Hence, we can solve tasks requiring both invariant and equivariant representations. A classic experiment in cognitive science concerns so-called {\it mental rotation}, i.e., the cognitive ability to manipulate mental representations of objects in 3D space. In the Shepard and Metzler paradigm, participants view two images of an unfamiliar 3D shape, where one image is a rotated, and possibly mirrored, version of the other, and must decide whether the two depict the same object or mirror-reflected objects~\cite{shepard1971mental}. Solving this \emph{mental rotation problem} requires equivariance, i.e., faithful representation of an object's ``pose" \cite{shepard1971mental,carroll1993human}. 
\begin{figure}[!b]
  \centering
  \vspace{-1em}
\includegraphics[width=.98\linewidth]{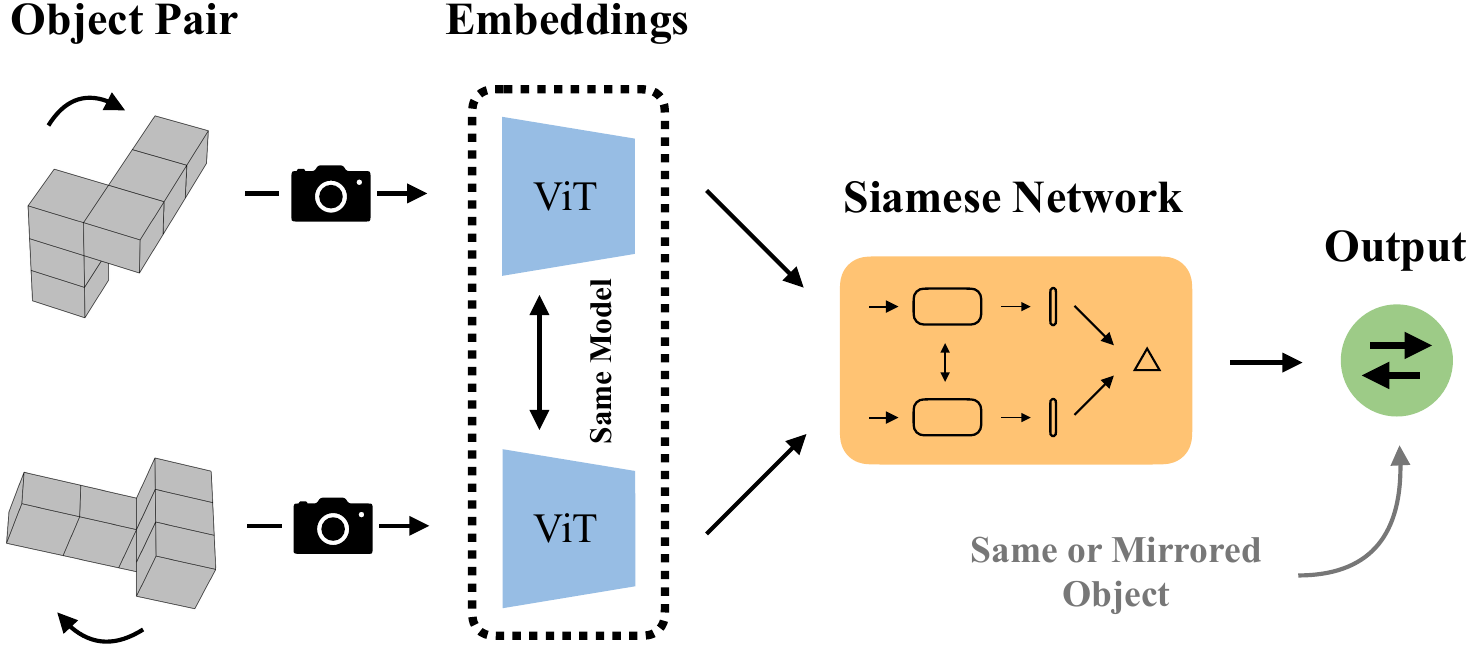}
  \caption{Overview of the experimental pipeline for the mental rotation task. Pairs of objects are rendered from different viewpoints and passed through a shared Vision Transformer (ViT). The resulting embeddings are compared in a Siamese network to determine whether the two views correspond to the same object under rotation or to a mirrored counterpart. This setup tests whether model representations preserve pose information sufficient to solve mental rotation problems.}
  \label{fig:figure_1}
\end{figure}
In computer vision, robustness to rotations is often pursued through \emph{rotation invariance} in image representations. Conventional convolutional neural networks are not inherently rotation invariant, so prior work embeds rotational structure into models using data augmentation with rotated images or specialized architectures such as group-equivariant CNNs and spatial transformer networks~\cite{cohen2016group,jaderberg2015spatial}. Most efforts, however, target in-plane image rotations, often at 90$^\circ$ increments or small angles, and do not address full 3D object rotations~\cite{esteves2018learning}. These methods are typically effective only near specific rotation angles and remain sensitive to arbitrary viewpoint changes~\cite{esteves2018learning,marcos2017rotation}. Moreover, a completely rotation-invariant embedding discards orientation information, which is undesirable for mental rotation, where the goal is to distinguish rotated identical objects from mirrored ones. Instead of total invariance, the representation should preserve pose cues sufficient to separate a rotated object from its mirror image.

Beyond task-level performance, recent studies have begun to open the ``black box" of Vision Transformers (\textit{ViT}) through layer-wise analyses. Such works show that different layers progressively specialize: early layers often capture low-level visual primitives (e.g., color, texture), while deeper layers encode more abstract semantic or relational features \cite{dorszewski2025colors,vielhaben2025beyond}. These analyses highlight that representations evolve throughout the network, and that intermediate layers can carry information that is discarded or abstracted away at the final layer. For our problem, this is crucial: mental rotation requires preservation of pose information that may be encoded more strongly at intermediate stages than in the final semantic embedding.

This work focuses on the mental rotation problem in the context of modern high-capacity vision models. The overview of our pipeline is illustrated in \autoref{fig:figure_1}. We investigate whether image embeddings from Google’s supervised Vision Transformer (\texttt{ViT})~\cite{dosovitskiy2021vit} and other large-scale ViT-based representation models trained with self-supervision, such as \texttt{CLIP}~\cite{radford2021clip}, \texttt{DINOv2}~\cite{oquab2023dinov2}, and \texttt{DINOv3}~\cite{siméoni2025dinov3}, implicitly encode 3D structural and viewpoint information sufficient to distinguish an object from its mirrored counterpart under different rotations.  In this paper, we use ViT to denote the Vision Transformer architecture in general, and \texttt{ViT} to refer specifically to Google’s supervised model.  \texttt{CLIP} learns transferable image features by aligning images with text descriptions, while \texttt{DINOv2} and \texttt{DINOv3} are self-supervised transformers that report strong general-purpose visual features. We ask whether embeddings from these models, both at the final layer and at intermediate layers, contain information that enables mental rotation-like reasoning. Evidence of such information would suggest a form of \emph{viewpoint equivariance} that preserves pose, which is appropriate for this task. By analyzing and comparing embeddings on image pairs depicting rotated or mirrored objects, we assess whether contemporary representations encode the necessary cues for mental rotation, or whether new architectures and training paradigms are required.

Specifically, we find that: i) self-supervised ViTs (\texttt{CLIP}, \texttt{DINOv2}, and \texttt{DINOv3}) capture geometric structure better than supervised ViTs; ii) intermediate layers outperform final layers; iii) task difficulty scales with rotation complexity and occlusion, paralleling human reaction times and suggesting analogous constraints in embedding space representations.

All code for generating the Shepard-Metzler and the text data, as well as the code for all our experiments, is available on GitHub\footnote{\url{https://github.com/gjoelbye/vit-mental-rotation}}.
\begin{figure}
  \centering
\includegraphics[width=\linewidth]{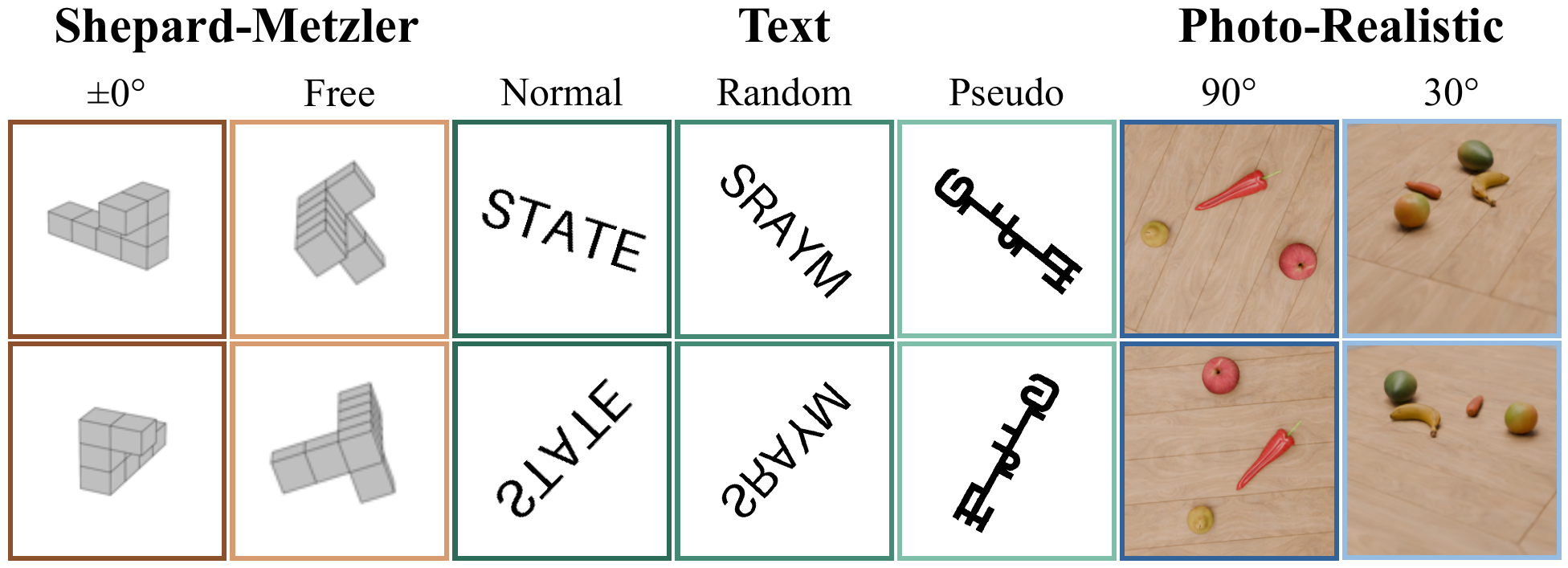}
 \vspace{-2.0em}
  \caption{Overview of the seven dataset variants used in the experiments. Shepard-Metzler tasks include pairs with small relative elevation rotations (±0°) and unconstrained rotations (Free). Text tasks include natural words (Normal), randomly sampled character strings (Random), and artificial symbols rendered in the PseudoSloan font (Pseudo). Photo-Realistic tasks consist of tabletop object scenes captured from two viewpoints, with either a 90° or 30° camera azimuth angle.} 
  \vspace{-1.0em}
  \label{fig:figure_2}
\end{figure}
\begin{figure*}[!t]
  \centering
  \includegraphics[width=1.01\textwidth]{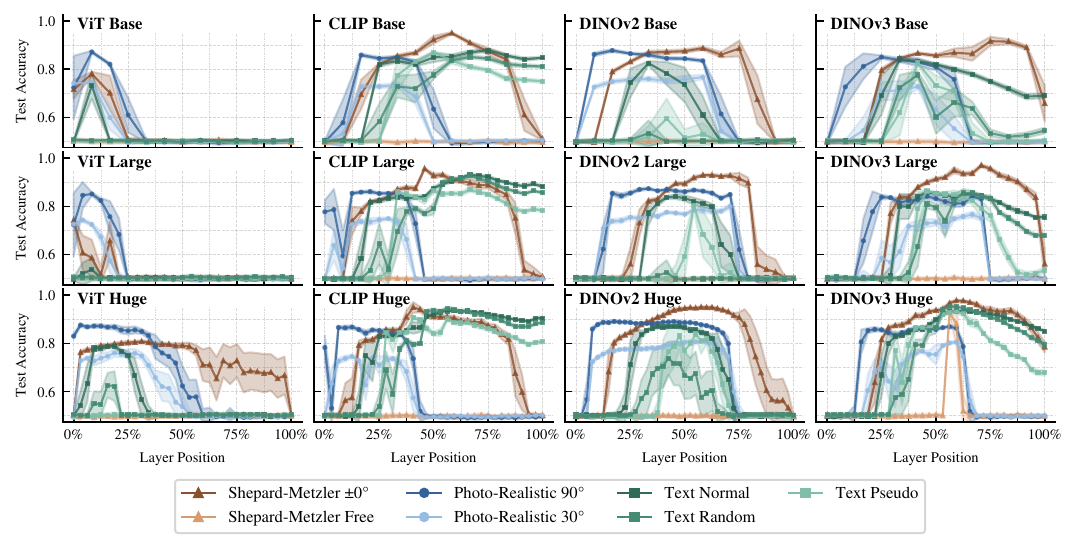}
  \vspace{-1.5em}
  \caption{Test accuracy across layers for four model families (\texttt{ViT}, \texttt{CLIP}, \texttt{DINOv2}, \texttt{DINOv3}) in three sizes (\texttt{Base}, \texttt{Large}, \texttt{Huge}). Results are averaged over three runs of stratified 10-fold cross-validation, with shaded regions indicating standard error. Text-related tasks are preserved in the final layers only for \texttt{CLIP} and \texttt{DINOv3}, whereas all models retain text-related information in earlier layers. Photo-Realistic scenes and Shepard-Metzler ±0° objects are consistently well-represented in early and middle layers, but most models lose this signal toward the final layers, except for \texttt{DINOv3}. For Shepard-Metzler Free objects, only the deepest layers of \texttt{DINOv3} \texttt{Huge} (layers 18–19) maintain discriminative information.}
  \vspace{-1em}
  \label{fig:figure_3}
\end{figure*}
\section{Data Generation}
To probe how vision models handle rotation and mirroring across different domains, we design three families of synthetic datasets: Shepard-Metzler objects, Text, and Photo-Realistic tabletop scenes. Examples are shown in \autoref{fig:figure_2}. Each dataset produces pairs of images where the positive label corresponds to the same object or scene under a 3D rotation, and the negative label corresponds to a mirrored counterpart. For each dataset, we generate $20{,}000$ balanced pairs.

\textbf{Shepard-Metzler.}  
We construct 3D block structures inspired by the classic Shepard-Metzler paradigm. Shapes consist of 5--9 unit cubes arranged in straight segments of length 2--4, with occasional orthogonal branching. Overlaps are disallowed, and the maximum branching degree is bounded. To ensure spatial complexity, only shapes that span all three axes and are not mirror-symmetric across the $x$-axis are retained. Each shape is rendered from two perspectives under random elevation and azimuth angles. We call this setting ``Free". An easier version of this task is restricting the elevation of the second view to lie within $\pm \theta_1$ of the first, e.g., Shepard-Metzler ±0° have different azimuth but the same elevation angle.  With probability $1/2$, the second projection mirrored. 

\textbf{Text.}  
We generate character strings of length 3--6 under three conditions. \emph{Normal} strings are sampled from the 10{,}000 most frequent English words. \emph{Random} strings are sampled uniformly from the alphabet. \emph{Pseudo} strings are sampled uniformly but rendered in the artificial \textit{PseudoSloan} font \cite{Vildavski2022PseudoSloan}. Each string is randomly rotated and flipped horizontally with probability $1/2$, such that half of the resulting pairs are rotationally equivalent and half are mirrored.

\textbf{Photo-Realistic.}  
To approximate more natural inputs, we use Blender\footnote{An open-source 3D rendering suite: \url{https://www.blender.org}} to create tabletop scenes with 3--6 pieces of fruit drawn with replacement from a pool of 20 items. Each object is placed at a random position and orientation. Scenes are photographed under perspective projection from two viewpoints at fixed elevation $\theta_2$ but random azimuth angles. With probability $1/2$, the second view is mirrored with respect to the table center. Note that only the items on the tabletop are mirrored and not the tabletop itself. These datasets should be more aligned with the photo-realistic prior of vision transformers while probing their sensitivity to viewpoint and mirroring. 

\section{Methods}
\textbf{Models.}
Our experiments are based on four pre-trained ViT encoders that have each seen broad adoption in literature. Although they share the same backbone, they differ in level of supervision, objective and datasets. Google's \texttt{ViT} \cite{radford2021clip} is pretrained with categorical supervision on the ImageNet-21K dataset (\(\sim \)4M images, 21,843 categories), using a cross-entropy classification objective. Images are tokenized into fixed-size patches, passed through a transformer, and optimized to predict the ground-truth label. The training recipe follows standard image classification practice, where images are augmented through random crops and horizontal flips \cite{steiner2022trainvitdataaugmentation}. As the target variable is class-identity only, the training loss therefore enforces features that are stable across within-class variations, such as viewpoint and small geometric transformations. 

OpenCLIP \cite{Cherti_2023} is a faithful open-source reproduction of OpenAI's \texttt{CLIP} models, trained on the LAION-2B dataset. For consistency, the name \texttt{CLIP} is used throughout this work. Regarded as a semi-supervised contrastive model, an image encoder (ViT) and a transformer-based text encoder are trained jointly through a symmetric InfoNCE objective, where matching images and captions are pulled together and mismatches are pushed apart. 
In contrast to the supervised \texttt{ViT}, the augmentation strategy is minimal. Typically limited to random resized crops, since supervision comes primarily from the paired text. As captions may explicitly describe orientation or viewpoint, the objective encourages the encoders to preserve visual cues useful for alignment, rather than enforcing invariance. 

Meta’s \texttt{DINOv2} \cite{oquab2023dinov2} and \texttt{DINOv3} \cite{siméoni2025dinov3} are self-supervised ViTs trained without labels on large-scale curated image datasets. \texttt{DINOv2} is trained on LVD-142M, a 142M–image collection automatically curated to resemble established smaller manually curated datasets. \texttt{DINOv3} scales the approach to frontier-size models with up to 7B parameters, trained on LVD-1689M, a broader 1.7B–image dataset constructed with the same principle but at a much larger scale.
Both models rely on a teacher–student framework where several  
augmented views of the same image are generated through random resized crops and standard augmentations such as horizontal flips, color jitter, blur, and solarization \cite{moutakanni2024dontneeddomainspecificdata}. The student processes all views, while the teacher, updated as an exponential moving average of the student, receives only the two global views. The teacher produces softmax distributions over a very large output space, and the student is trained with a cross-entropy loss to match these outputs, thereby enforcing augmentation-invariant representations without reconstruction. The models rely on distribution matching through image-level, patch-level, and uniformity losses. In \texttt{DINOv2}, this serves as the central training objective, whereas \texttt{DINOv3} extends the framework with so-called Gram anchoring, a new loss term that stabilizes dense features in large-scale training \cite{siméoni2025dinov3}. This extension enables \texttt{DINOv3} to retain the robust global representations of \texttt{DINOv2} while further improving the stability of high-resolution dense maps. 

We use three sizes of each architecture: \texttt{Base}, \texttt{Large}, and \texttt{Huge} (\texttt{Giant} in \texttt{DINOv2}-paper). For \texttt{ViT} and \texttt{CLIP}, all variants are trained from scratch on their respective datasets, whereas for \texttt{DINOv2} and \texttt{DINOv3}, only the largest model is trained from scratch, with \texttt{Base} and \texttt{Large} obtained through distillation.\\
\textbf{Experiment.}
For each dataset of 20\,000 pairs, we extract the output of every transformer layer from each model and compute the mean over the patch tokens. This produces an embedding vector for each layer in each model, with dimensionalities of 768, 1024, and 1280 for the \texttt{Base}, \texttt{Large}, and \texttt{Huge} variants, respectively. Thus, we obtain 20\,000 embedding pairs per layer for every model.

We train a lightweight Siamese classifier on these embedding pairs to assess whether the encoder representations support mental rotation. A pair $\mathbf{z}_1,\mathbf{z}_2\in \mathbb{R}^d$ are passed through a shared MLP $f_\theta:\mathbb{R}^d\rightarrow\mathbb{R}^{128}$ with batch normalization and ReLU activations, then $\ell_2$-normalized to $\mathbf{\tilde{z}}_i = f_\theta(\mathbf{z}_i)/\lVert f_\theta(\mathbf{z}_i)\rVert_2$. The interaction vector is the elementwise absolute difference $\mathbf{\Delta \tilde{z}} = \lvert \mathbf{\tilde{z}}_1 - \mathbf{\tilde{z}}_2 \rvert$, which is fed into a logistic head $h_\phi$ to compute $p(y=1\mid \mathbf{z}_1,\mathbf{z}_2) = \sigma(h_\phi(\mathbf{\Delta \tilde{z}}))$. A positive label indicates that the two images show the same instance under rotation, while a negative label corresponds to a mirrored version. We train using binary cross entropy, AdamW (learning rate $10^{-3}$), a batch size of 256, and a learning rate schedule that linearly warms up for 15 epochs followed by cosine annealing, up to a maximum of 200 epochs. Early stopping monitors validation loss with a patience of 50 epochs. Before training, we standardize each embedding using the mean and standard deviation computed from the training split only, pooling across both elements in each pair. We perform three-times repeated stratified 10-fold cross-validation, reserving 10\% of each training fold for validation to select the epoch. We report accuracy and cross entropy on the held-out test fold as mean $\pm$ standard error across folds. This protocol is repeated independently for each layer and each dataset to isolate their effects. Overall, signal strength varies across embeddings, which motivates the extensive training procedure to ensure reliable signal capture.
\begin{figure}[!t]
  \centering
  \includegraphics[width=\linewidth]{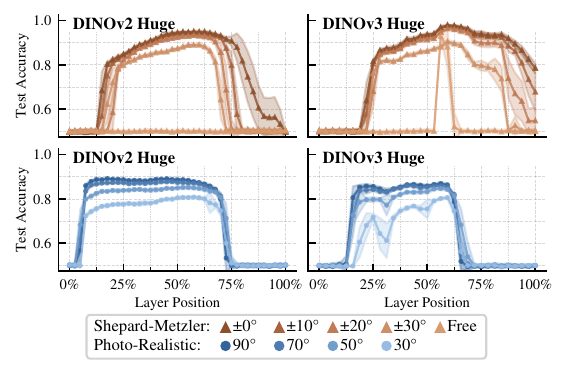}
  \vspace{-2em}
  \caption{Test accuracy across layers for \texttt{DINOv2} \texttt{Huge} and \texttt{DINOv3} \texttt{Huge} on extended Shepard-Metzler (top) and Photo-Realistic (bottom) datasets. Shepard-Metzler tasks vary in relative rotation angle (±0° to Free), and Photo-Realistic tasks in camera elevation (30°–90°). Accuracy declines as transformations become harder, with both models strongest in early/mid layers for simple rotations and weakest for large pose changes or unconstrained rotations.
  }
  \vspace{-1em}
  \label{fig:figure_4}
\end{figure}

\section{Results and discussion}
We show the development of the models' capability to solve the variants of the mental rotation problem in \autoref{fig:figure_3}. In \autoref{fig:figure_4}, we show how the \texttt{Huge} variants of \texttt{DINOv2} and \texttt{DINOv3} handle the gradually increasing difficulty of the Shepard-Metzler and Photo-Realistic tasks.

Across all tasks, \texttt{CLIP} consistently demonstrates strong performance, particularly in scenarios involving textual information. This is unsurprising given \texttt{CLIP}’s joint training on text–image pairs, which naturally gives it OCR-like capabilities \cite{radford2021clip}. What is especially striking, however, is that \texttt{DINOv3} performs comparably well in these text-heavy conditions, despite not being explicitly trained with text supervision. In contrast, supervised ViT models perform poorly relative to their self-supervised counterparts. A possible explanation is that supervised ViTs, optimized for categorical discrimination, favor invariance over pose sensitivity, thereby discarding geometric information needed for mental-rotation tasks.
For the simplest Shepard-Metzler ±0° problem, all models reach non-trivial accuracy at some point in the network. The stage at which performance peaks, however, differs: \texttt{ViT} \texttt{Base} and \texttt{ViT} \texttt{Large} exhibit the strongest representations early in the network, \texttt{ViT} \texttt{Huge} shows weak but consistent performance throughout, and the rest of the models peak at intermediate depths. The difficulty of the tasks increases with increasing the relative rotational angle. This parallels human reaction times in mental rotation \cite{shepard1971mental}, and is consistent with findings that humans and deep networks face similar difficulties with geometric transformations \cite{kheradpisheh2016humans}. Fruit rotations show a similar pattern, with accuracy declining as object occlusion increases.

The Shepard-Metzler Free condition represents the most challenging setting. Strikingly, only \texttt{DINOv3} \texttt{Huge} shows any ability to solve this task, and even then, only at layer 18.
A recurring pattern across models is the emergence of a plateau in task performance in mid-level layers, followed by a decline toward the final layers. This aligns with findings that different types of concepts are encoded at different depths of the transformer hierarchy \cite{dorszewski2025colors, vielhaben2025beyond}, and echoes observations across vision, speech, and language models that intermediate representations often provide richer or more useful structure than the final layer (e.g. \cite{tetkova2025convex, dorszewski2025connecting, skean2025layer}). In practical terms, this indicates that the final representation layer is not always the most informative for tasks requiring geometric reasoning, and that intermediate representations may better capture the relevant structure.

Using the CLS token instead of patch-averaged embeddings significantly reduces performance across models (data not shown), although the qualitative trends remain the same. This finding aligns with prior work suggesting that CLS encodings are less robust than spatially averaged patch embeddings for non-classification tasks \cite{chu2023conditional}.

\begin{figure}[!t]
  \centering
  \includegraphics[width=\linewidth]{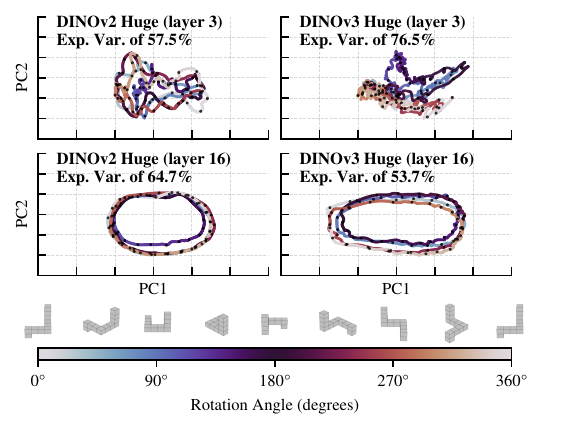}
  \vspace{-2.5em}
  \caption{Projection onto the first two principal components of \texttt{DINOv2} \texttt{Huge} and \texttt{DINOv3} \texttt{Huge} for layer 3 (when the model cannot yet solve the task) and layer 16 (when it can). For the simple Shepard-Metzler task (±0°) with blocks rotated incrementally over 360°, the later layer shows a clear, continuous structure aligned with rotation angle, unlike the early layer.}
  \vspace{-0.5em}
  \label{fig:figure_5}
\end{figure}

In \autoref{fig:figure_5}, we inspect the principal component space of the best- and worst-performing layers of the \texttt{DINOv2} \texttt{Huge} and \texttt{DINOv3} \texttt{Huge} models for a Shepard-Metzler object rotated 360°. In the high-performing layers, the embeddings form a smooth, circular structure aligned with the rotation angle, highlighting the emergence of equivariant behavior in contrast to earlier layers where this structure is absent.

We also evaluated Meta's \texttt{MAE} ViT \cite{maevit} in the \texttt{Base}, \texttt{Large}, and \texttt{Huge} variants. Unlike contrastive or teacher–student approaches, this masked autoencoder model failed to solve the mental rotation problem at any layer, suggesting that reconstruction-based self-supervision does not capture the geometric structure required for pose-sensitive reasoning.

\section{Conclusion}Our study shows that large vision models can solve mental rotation tasks, but performance depends on architecture, supervision, and depth. Self-supervised transformers outperform supervised ViTs, suggesting that objectives encouraging geometric sensitivity better support spatial reasoning. Pose information is strongest in intermediate layers but often lost in final embeddings. Task difficulty scales with rotation angle and occlusion, mirroring human performance. These findings highlight both the promise of current models and the need for approaches that preserve geometric structure more faithfully across layers.

\clearpage
\section{Acknowledgements}
This work was supported by the Novo Nordisk Foundation grant NNF22OC0076907 "Cognitive spaces - Next generation explainability" and the Pioneer Centre for AI, DNRF grant number P1.
It was also supported by the Danish Data Science Academy, which is funded by the Novo Nordisk Foundation (NNF21SA0069429) and VILLUM FONDEN (40516).

\bibliographystyle{IEEEbib}
\bibliography{IEEEbib}

\end{document}